\documentclass[10pt,journal,compsoc]{IEEEtran}

\usepackage{cite}
\usepackage{url}
\usepackage{ragged2e}
\usepackage{footnote}
\makesavenoteenv{tabular}
\makesavenoteenv{table}
\usepackage{epsfig}
\usepackage{graphicx}
\usepackage{amsmath,amssymb} %
\usepackage{algorithm}
\usepackage{algorithmicx}
\usepackage{algpseudocode}
\usepackage{graphics}
\usepackage{threeparttable}
\usepackage{color}
\usepackage[normalem]{ulem}
\usepackage{multirow}
\usepackage{float}
\usepackage{amsfonts}
\usepackage{bm}
\usepackage{array}
\usepackage{pifont}
\usepackage{diagbox}
\usepackage{rotating}
\usepackage{booktabs}
\usepackage{overpic}
\usepackage{textcomp}
\usepackage{contour}
\usepackage{makecell}

\usepackage{enumitem}
\usepackage{colortbl}
\usepackage{bbding}
\usepackage[breaklinks=true,colorlinks]{hyperref}

\usepackage{silence}
\newcommand{\figref}[1]{Fig.~\ref{#1}}
\newcommand{\tabref}[1]{Table~\ref{#1}}
\newcommand{\equref}[1]{Eq.~(\ref{#1})}
\newcommand{\secref}[1]{$\S$\ref{#1}}

\graphicspath{{./Imgs/},{./Imgs/Samples/},{./Imgs/ErrorAnalysis/},{./Imgs/Visualization/}}
\DeclareGraphicsExtensions{.pdf,.jpg,.png}

\def\sArt{{state-of-the-art~}}

\def\ie{\emph{i.e.}}

\def\etal{{\em et al.~}}
\def\sArt{{state-of-the-art~}}

\newcommand{\highlight}[1]{{\textbf{\textcolor{blue}{#1}}}}

\newcommand{\myPara}[1]{\vspace{.12in}\noindent\textbf{#1}}

\begin{document}

\title{Ret3D: Rethinking Object Relations for Efficient 3D Object Detection in Driving Scenes}

\markboth{IEEE TRANSACTIONS ON PATTERN ANALYSIS AND MACHINE INTELLIGENCE}%
{W\MakeLowercase{u} \MakeLowercase{\textit{et al.}}}

\author{
Yu-Huan Wu, Da Zhang, Le Zhang, Xin Zhan, Dengxin Dai, Yun Liu, and Ming-Ming Cheng
\IEEEcompsocitemizethanks{
    \IEEEcompsocthanksitem Y.-H. Wu and M.-M. Cheng are with TMCC, 
      College of Computer Science, Nankai University, Tianjin, China. 
      (wuyuhuan@mail.nankai.edu.cn, cmm@nankai.edu.cn)
    \IEEEcompsocthanksitem D. Zhang and X. Zhan are with Alibaba DAMO Academy. 
      (dazhang@cs.ucsb.edu) 
    \IEEEcompsocthanksitem L. Zhang is with University of Electronic Science 
      and Technology of China. (zhangleuestc@gmail.com) 
    \IEEEcompsocthanksitem D. Dai is with Max Planck Institute for Informatics, Germany. (ddai@mpi-inf.mpg.de)
    \IEEEcompsocthanksitem Y. Liu is with Computer Vision Lab, ETH Zurich, Switzerland. (yun.liu@vision.ee.ethz.ch)
    \IEEEcompsocthanksitem This work was done when Y.-H. Wu is a research intern at Alibaba DAMO Academy.
    \IEEEcompsocthanksitem Corresponding author: M.-M. Cheng (cmm@nankai.edu.cn).
}
}

\IEEEtitleabstractindextext{%
\begin{abstract}
  \justifying
Current efficient LiDAR-based detection frameworks are lacking in exploiting 
object relations, 
which naturally present in both spatial and temporal manners.
To this end, we introduce a simple, efficient, 
and effective two-stage detector, termed as Ret3D.  
At the core of Ret3D is the utilization of novel intra-frame and inter-frame 
relation modules to capture the spatial and temporal relations accordingly.
More Specifically, intra-frame relation module (IntraRM) encapsulates the 
intra-frame objects into a sparse graph and thus 
allows us to refine the object features through efficient message passing.
On the other hand, inter-frame relation module (InterRM) densely connects 
each object in its corresponding tracked sequences dynamically, 
and leverages such temporal information to further enhance its representations 
efficiently through a lightweight transformer network. 
We instantiate our novel designs of  IntraRM and InterRM with 
general center-based or anchor-based detectors and evaluate them 
on  Waymo Open Dataset (WOD). 
With negligible extra overhead, Ret3D achieves the \sArt performance,  
being 5.5\% and 3.2\% higher than the recent competitor in terms of the 
LEVEL\_1 and LEVEL\_2 mAPH metrics on vehicle detection, respectively.
\end{abstract}
\begin{IEEEkeywords}3D Object Detection, Object Relations, Autonomous Driving.
\end{IEEEkeywords}
}

\maketitle

\IEEEpeerreviewmaketitle

\section{Introduction}\label{sec:intro}
3D object detection aims at recognizing vehicles, pedestrians, cyclists, 
and other key features in large-scale scenes, 
and it is considered as one of the key components
of the perception system for autonomous driving \cite{sun2020scalability}.
In the course of the development of 3D object detection for autonomous driving,
LiDAR-based approaches \cite{yan2018second,lang2019pointpillars,shi2020pv,yin2021center} show
its superiority over monocular or multi-view image-based methods \cite{simonelli2019disentangling,chen2020monopair,reading2021categorical} because LiDAR signals provide accurate depth information via point clouds, even with a very long range, while cameras are naturally limited by purely 2D views.

Object relations have long proven to be profitable for 2D object detection \cite{tu2008auto,galleguillos2010context,mottaghi2014role,hu2018relation}.
Considering 3D object detection in autonomous driving,
the locations and geometric features of 3D objects can provide rich contextual and structural information for scene understanding and accurate object recognition.
However, current LiDAR-based detection frameworks are limited in exploiting object relations.
Specifically, most works only \textit{implicitly} explore object relations through the hidden features of the carefully-designed convolutional neural networks (CNNs) 
\cite{bewley2020range, miao2021pvgnet, shi2020pv} or vision transformers \cite{mao2021voxel}.
Some works \cite{qi2021offboard} leverage long point cloud sequences 
to improve the off-board detection while introducing much higher computational cost.
To the best of our knowledge, there are no principled solutions on how to \textit{explicitly} and efficiently leverage object relations for improving 3D object detection.

\newcommand{\AddImg}[1]{%
\includegraphics[width=0.48\columnwidth, height=0.6\columnwidth]{Imgs/#1}%
}

\begin{figure}[!t]
    \centering
    \footnotesize
    \renewcommand{\arraystretch}{0.6}
    \setlength{\tabcolsep}{0.35mm}
    \begin{tabular}{cc}
        \AddImg{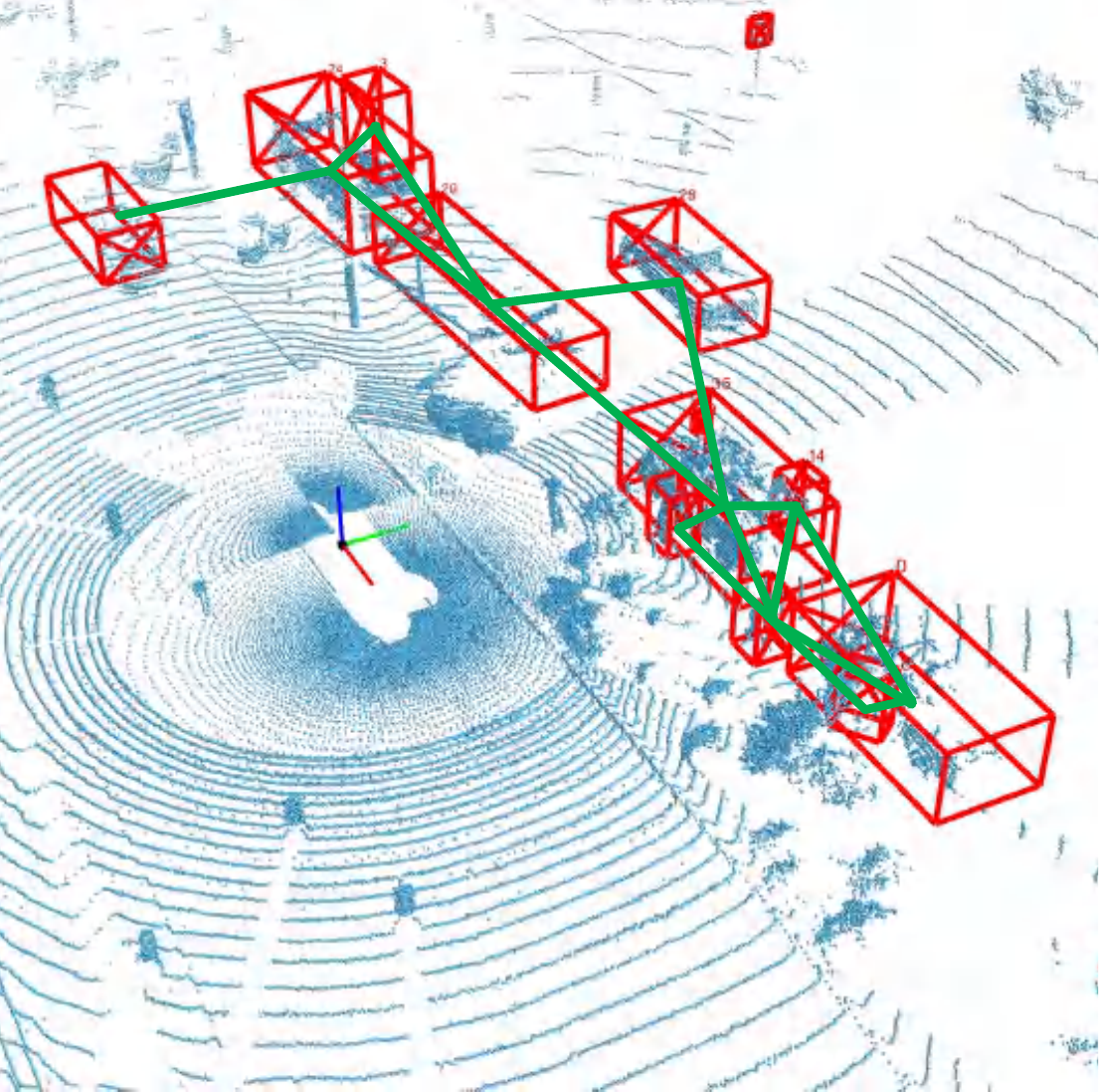}
        &
        \begin{overpic}[width=0.48\columnwidth, height=0.6\columnwidth]{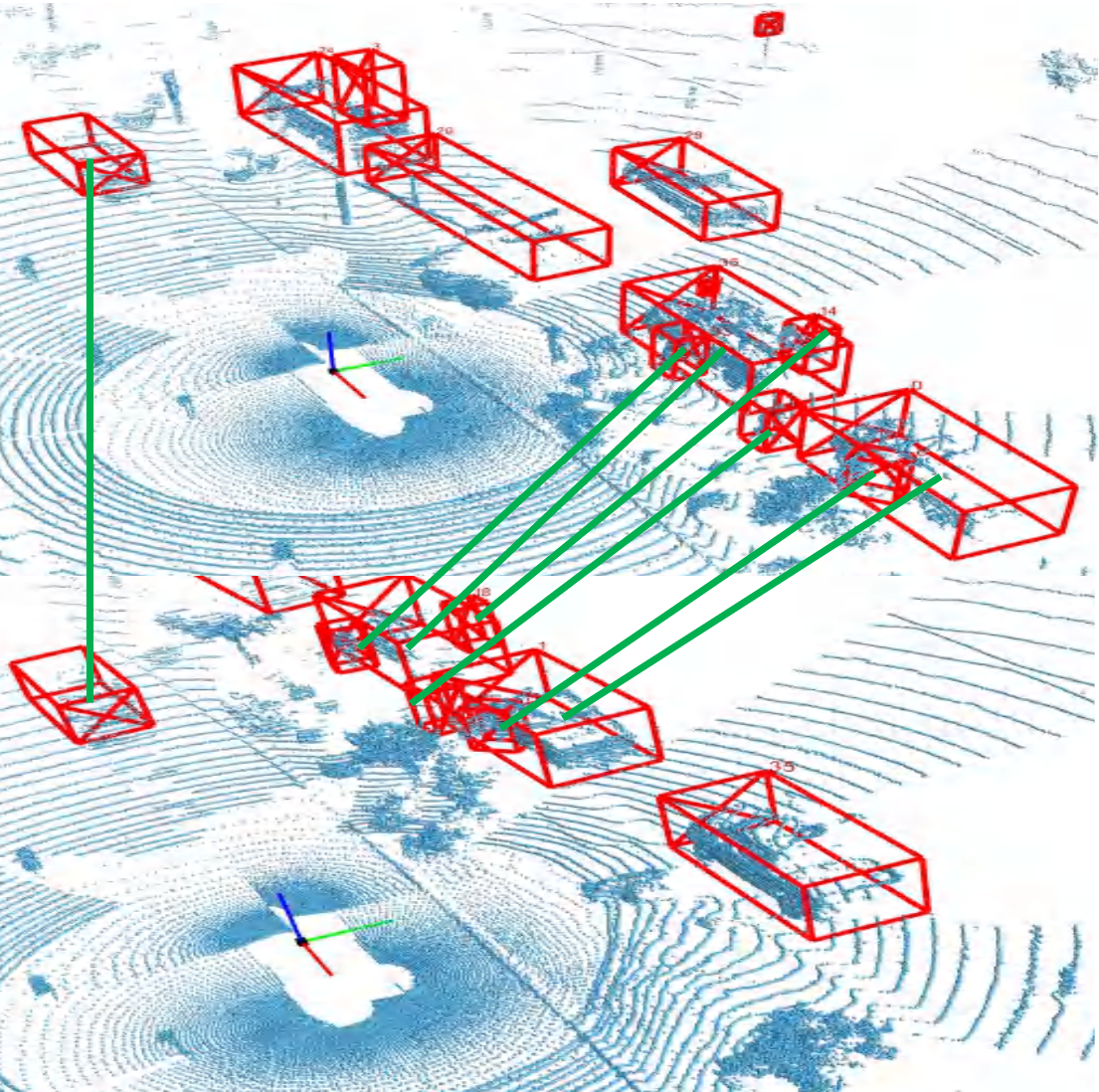}
        \put(65,93){$t=1$}
        \put(65,42){$t=2$}
        \end{overpic} 
        \\
        (a) Intra-frame object relations & (b) Inter-frame object relations
     \end{tabular}
    \vspace{-5pt}
    \caption{\textbf{An example of intra-frame and inter-frame object relations.}
    \textbf{\textcolor[rgb]{0.3, 0.68, 0.36}{Green}} lines indicate object relations. For simplicity, 
    only a short sequence with a length of 2
    is used to illustrate inter-frame object relations.
    Best viewed in color.}
    \label{fig:relation_example}
\end{figure}%

In practice, LiDAR frames come naturally in time as a sequence.
We consider two types of relationship: intra-frame and inter-frame object relations.
We define that
\textit{intra-frame} object relations are the relations of 3D objects within the current frame. 
In contrast, \textit{inter-frame} object relations indicate the spatial and temporal relations of the same 3D object across different frames in a long LiDAR sequence. 
For a better understanding, we provide an example for these two types of object relations in \figref{fig:relation_example}.
We believe that both \textit{intra-frame} and \textit{inter-frame} object relations are beneficial for 3D object detection.

For the \textit{intra-frame} object relation modeling, one na\"ive solution is to densely connect every object with all other objects in the same frame.
Although this can improve the detection results,
such a densely-connected graph contains lots of redundant information and inevitably induces more computational cost.
For example, a vehicle is rarely related to a pedestrian tens of meters apart.
Therefore, building a sparse graph via practical priors, such as the object's physical locations,  
can avoid redundant information and thus improve the efficiency. For the \textit{inter-frame} object relations, each object may appear in a long sequence.
It is again time-consuming to process such a long sequence by na\"ively aggregating point clouds and detection results \cite{qi2021offboard}.
Instead, we propose to leverage object-level information, \ie, location, heading, and velocity, for modeling inter-frame object relations, which largely alleviates the computational cost and thus keeps high efficiency for 3D object detection.

Based on the above observations, we introduce a spatial-temporal framework called Ret3D for two-stage 3D object detection.
In order to build intra-frame object relations,
we construct a sparse undirected graph. In this graph, each object is viewed as a node, and the spatial distance between two nodes determines if the corresponding edge exists.
Such a sparse graph allows us to iteratively refine 
the object relations and features through efficient message passing.
Here, only negligible computational cost ($<$0.1\% of the base detector) is needed due to the sparsity of the graph and the low dimensionality of object features in each frame.
Finally, object locations can be amended by the refined object features.
For the inter-frame object relations,
we introduce a transformer-based detector,
which efficiently models the densely-connected inter-frame relationship for each object through the tracked sequences.
In this way, our method can leverage the detection results of different time stamps.

Following \cite{deng2020voxel,yin2021center},
we conduct extensive experiments on the popular Waymo Open Dataset (WOD) \cite{sun2020scalability} to validate the proposed framework.
Experimental results demonstrate that both intra-frame and inter-frame object relations are significant for improving detection performance.
With CenterPoint~\cite{yin2021center} as the base detector,
the proposed method performs significantly better than recent \sArt methods under both the LEVEL\_1 and LEVEL\_2 settings \cite{sun2020scalability}.
Therefore, our idea about intra-frame and inter-frame object relation modeling opens
a new path for 3D object detection and would be useful for future research.

Overall, our contributions can be summarized as below:
\begin{itemize}
    \item We propose to explicitly learn intra-frame and inter-frame object relations for improving the accuracy of efficient 3D object detection in autonomous driving.
    \item We propose the intra-frame relation module (IntraRM), which constructs a sparse object graph to refine the feature of objects within the same LiDAR frame through message passing.
    \item We propose the inter-frame relation module (InterRM), which densely connects each object in its tracked sequences 
    to further refine its representations through a lightweight transformer network.
\end{itemize}

\section{Related Work}
\subsection{LiDAR-based 3D Object Detection}
3D object detection is the core problem in the perception system of autonomous driving \cite{geiger2012we,liang2019multi,yang20203dssd}.
Unlike detecting 2D objects that only have 4 degree-of-freedom (DoF),
typical 3D objects have at least 7 DoF with 3D center location (x, y, z), length, width, height, and heading degree.
Therefore, 3D object detection requires much sensitive depth prior, which could be easily obtained from LiDAR signals. Hence, LiDAR-based 3D object detection \cite{yan2018second,lang2019pointpillars,shi2020pv,yin2021center} achieves better performance compared to others \cite{simonelli2019disentangling,chen2020monopair,reading2021categorical} in recent years.

Typically, for autonomous driving, each LiDAR frame often has hundreds of thousands of points \cite{geiger2012we,sun2020scalability}.  
Directly searching 3D objects in the point clouds is too 
challenging due to point clouds' sparsity,
irregularity, and disorder.
The common idea is to transfer irregular point clouds to features with regular grids.
For example, Vote3Deep \cite{engelcke2017vote3deep} transfers the point clouds to regular voxels via feature-centric voting with real-time speed.
VoxelNet \cite{zhou2018voxelnet} further proposes to 
derive more comprehensive voxels via PointNet \cite{qi2017pointnet} and extracts features with 3D sparse convolutions.
SECOND \cite{yan2018second} then
 largely speeds up 3D sparse convolution for better efficiency.
Some other works \cite{yang2018pixor,lang2019pointpillars} compute pillar features 
using the fixed encoding or PointNet \cite{qi2017pointnet}
from the point clouds in the bird's eye view (BEV),
then utilizing faster 2D CNN for feature extraction.
Inspired by \cite{duan2019centernet,tian2020fcos},
PillarOD \cite{wang2020pillar} and CenterPoint \cite{yin2021center} introduce anchor-free 3D object detection frameworks,
which replace conventional anchor-based prediction with pillar-centric anchor-free prediction.

\begin{figure*}
    \centering
    \includegraphics[width=.9\textwidth]{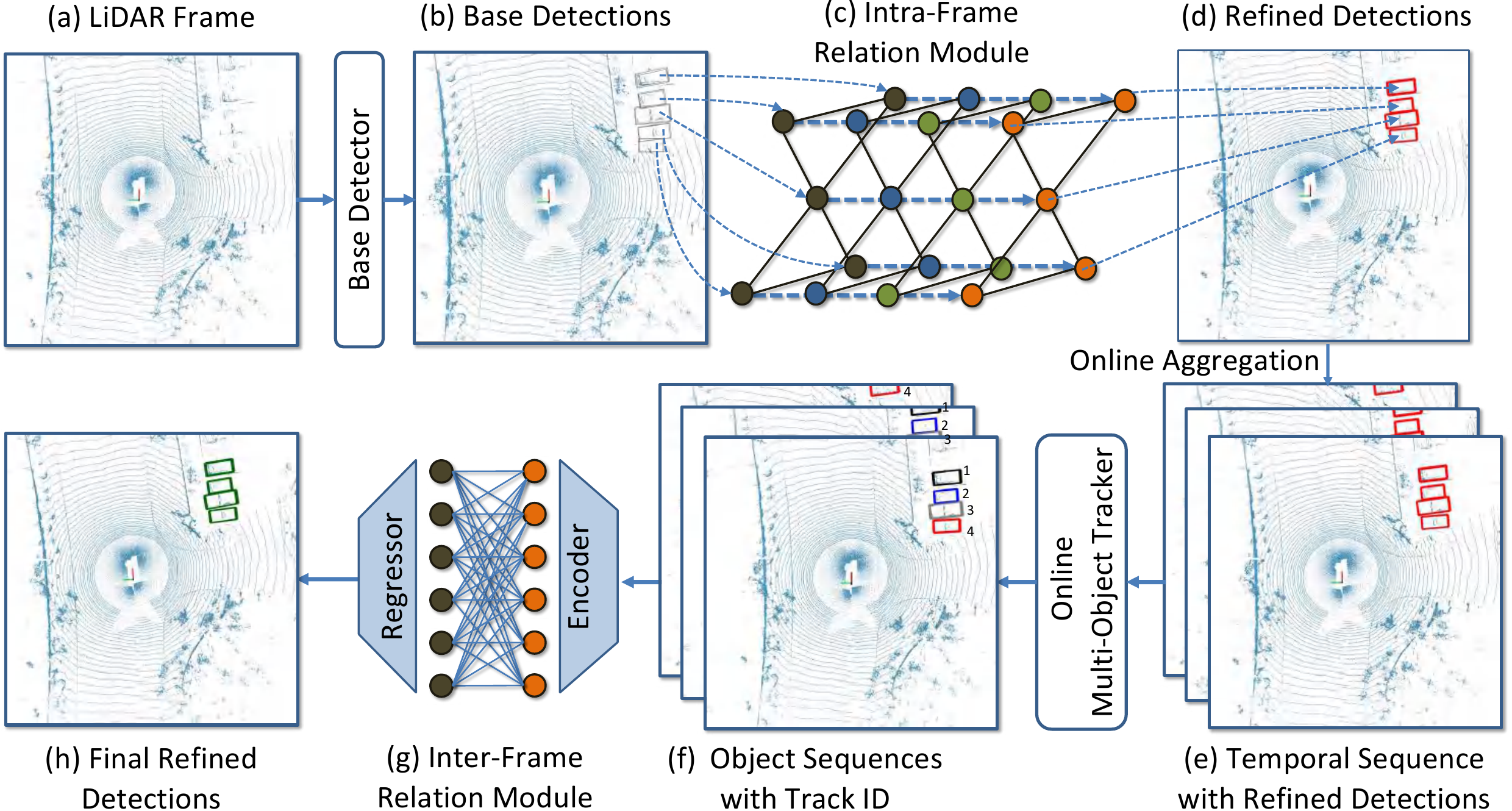}
    \vspace{-10pt}
    \caption{\textbf{The pipeline of Ret3D.} Ret3D is a two-stage detector that refines the detection results of one-stage detectors efficiently. 
    Ret3D consists of two parts, IntraRM and InterRM, for refining detection results using intra-frame and inter-frame object relations, respectively.
    }
    \label{fig:pipeline}
\end{figure*}

\subsection{Two-stage 3D Object Detection}
Recently, two-stage 3D object detectors become popular 
due to their strong compatibility and refinement given the regions of interest (RoI).
Many approaches \cite{chen2019fast,shi2019pointrcnn,shi2020pv,shi2020points,yin2021center,li2021lidar,deng2020voxel} 
adapt 2D R-CNN style frameworks \cite{ren2016faster} to 3D object detection in the BEV domain.
First, object proposals are generated by an RPN. 
An additional regression head 
is then used to score and rectify each object proposal independently.

Nevertheless, directly applying the above strategy is suboptimal as BEV features are also sparse,
and information loss is inevitable during this process.
Therefore, many approaches
\cite{yang2019std,shi2019pointrcnn,chen2019fast,shi2020pv,sheng2021improving,li2021lidar} work on leveraging point or voxel features, deriving more abundant spatial information.
For example,
Point R-CNN \cite{shi2019pointrcnn} leverages original point features in each RoI.
PV-RCNN \cite{shi2020pv} further proposes point-voxel set abstraction, 
encoding rich point and voxel features in each RoI.
LiDAR R-CNN \cite{li2021lidar} introduces a simple strategy for refining the detection results, which feeds the point cloud inside and around each detection to PointNet \cite{qi2017pointnet}.
Two-stage CenterPoint \cite{yin2021center} also refines the results but only uses 5-point BEV features instead of original point features in each object.

\subsection{Graph Networks for 3D Perception}
Due to the effectiveness of extracting geometric features on the graph,
graph networks are very popular for 3D perception.
Several works leverage graph networks to further strengthen the extracted 
point \cite{wang2019dynamic, najibi2020dops}, voxel \cite{he2020svga}, region \cite{feng2021relation}, or BEV \cite{wang2021object} features.
For example, Wang \etal \cite{wang2019dynamic} proposed the dynamic graph CNN (DGCNN) for indoor 3D analysis, in which each point is viewed as a node and the graph is updated dynamically for each graph iteration on the graph.
Shi \etal \cite{shi2020point} proposed Point GNN, which encodes the initial point clouds to a large sparse graph with auto-registration mechanism to simultaneously detect multiple objects.
Wang \etal \cite{wang2021object} leveraged query-based GNNs to further strengthen the sparse BEV features.

From another perspective, we propose IntraRM based on the graph to model the object relations within the same LiDAR frame.
IntraRM explicitly learns object relation features using GNNs and refines the detection result of each object.
The proposed IntraRM is computationally efficient 
due to the simple representation of object features and the sparsity of the object relation graph.
IntraRM can be easily plugged into modern efficient 3D object detectors.

\subsection{Transformer for 3D Object Detection}
Transformer is originally the dominating tool for natural language processing (NLP) since it is reliable for catching long-range relationships via multi-head self-attention (MHSA).
As global relationship is also essential for vision tasks,
Carion \etal~\cite{carion2020end} proposed DETR, adapting transformer to 2D object detection and largely simplifying the detection pipeline.
Inspired by DETR \cite{carion2020end}, 
many works \cite{dosovitskiy2021image, liu2021swin, wang2021pyramid, chen2021pre, wu2021p2t, liu2021transformer} adapt transformer to vision tasks and show transformer can surpass CNNs \cite{he2016deep, huang2017densely} on most vision tasks, such as image  classification \cite{dosovitskiy2021image}, object detection \cite{zhu2020deformable}, and 
semantic segmentation \cite{chen2021pre}.
Like the above significant progress,
recently, transformer has also achieved great success on 3D object detection.
For example,
voxel transformer \cite{mao2021voxel} introduced a 3D sparse transformer for voxel feature extraction, replacing the conventional 3D sparse CNNs.
CT3D \cite{sheng2021improving} extracts point features in each proposal and then individually leverages channel-wise transformer for proposal refinement.
Based on PointNet++ \cite{qi2017pointnet},  
Liu \etal \cite{liu2021group} introduced a group-free framework via transformers for set-to-set box prediction like DETR \cite{carion2020end}.

Unlike other approaches,
we propose the inter-frame relation module (InterRM), which efficiently extracts inter-frame object relations using transformers with tracked object sequences.
Each detected object is refined by strong priors of its detected history locations, sizes, and motions.

\section{Methodology}

In this section, we introduce the overall pipeline of our Ret3D in \secref{sec:pipeline}. 
Ret3D consists of two efficient modules, 
intra-frame relation module (IntraRM) and inter-frame relation module (InterRM) 
which are introduced in detail in \secref{sec:intrarm} and \secref{sec:interrm}, respectively. Finally, the analyses about the time complexity are provided in \secref{complexity} to show the efficiency of the proposed method.

\subsection{Pipeline}
\label{sec:pipeline}

We illustrate the pipeline of Ret3D in \figref{fig:pipeline}.
Given the point cloud $\mathcal{P}=\{p_1, p_2, ..., p_M\}$ containing $M$ points,
a one-stage base detector transforms the point clouds to regular voxels or pillars for further processing. 
Without loss of generalizability and for simplicity, we assume that point clouds are transformed to regular voxels.

\myPara{One-stage base detector.}
First, a 3D backbone network can be utilized to extract regular map-view features $\mathcal{B} \in \mathbb{R}^{C\times H \times W}$ from the voxels, where 
$H$ and $W$ are determined by the initial voxel size and the stride of the 3D backbone, 
and $C$ is the number of channels of map-view features $\mathcal{B}$.
After deriving the map-view features $\mathcal{B}$, a regression head predicts a set of detection results $\mathcal{D}$, 
which contain the center, size, heading, and velocity of each detected object. 
We define the detection results $\mathcal{D}$ as the basic features for each detected object.
Meanwhile, 
according to the center locations of all detection results,
we crop the feature vectors $\mathcal{O}$ on the map-view features $\mathcal{B}$.

\myPara{IntraRM.}
To model intra-frame object relations, 
we propose IntraRM, as shown in \figref{fig:pipeline} (c).
In this module,
we collect all basic features from the one-stage base detector. 
Additionally, for effectively modeling each object, we also extract each object's BEV feature vector 
whose location on the BEV features corresponds to the center of each detected object.
A sparse graph is then constructed according to the spatial location of each object.
Then, we refine the features of each object by message passing on the graph, 
upon which we could further obtain the refined detection results. 
For more details, please refer to \secref{sec:intrarm}.

\myPara{InterRM.}
From another perspective, as illustrated in \figref{fig:pipeline} (g), we propose InterRM to model inter-frame object relations,
which refines detection results with tracked object sequences. For the consistency with real-world applications, we only utilize preceding frames while future frames are ignored.
Given the object sequences, we extract features using the vision transformer, 
which is efficient and powerful for capturing global relationships, and regress the refined detection result for each object.
More details will be introduced in \secref{sec:interrm}.

\subsection{Intra-Frame Relation Module}\label{sec:intrarm}
Existing two-stage approaches only refine detection results individually 
via point \cite{shi2020pv, sheng2021improving}, voxel \cite{he2020svga}, or BEV \cite{yin2021center} features.
They do not consider the relationship between each object.
We argue that intra-frame object relations for detection refinement are also very significant.
Therefore,
we propose an efficient solution with a new intra-frame relation module (IntraRM).
With IntraRM, objects are connected as a sparse graph since a dense graph contains lots of redundant information with much larger computational cost.
On the other hand,
a sparse graph only contains important edges that can allow us to refine each object via efficient message passing.
The architecture is illustrated in \figref{fig:pipeline} (b) - (d).

For each LiDAR frame input, the base detector has detected $n$  objects 
with basic features $\mathcal{D}=\{d_1, d_2, ..., d_n\}$ and corresponding map-view feature vectors $\mathcal{O}=\{o_1, o_2, ..., o_n\}$.
A sparse graph $G = (V, E)$ is constructed on these $n$ detected objects, 
where the node set $V = \{v_1, v_2, ..., v_n\}$ contains all detected objects, 
and the edge set $E$ connects the node pairs.

Denoting the initial features of nodes $v_i, v_j$ as $x_{i}^0$, $x_{j}^0$,
they can be formulated as:
\begin{equation}
    x_{i}^0 = {\rm Concat}(d_i, o_i)\mathbf{W}^x, x_{j}^0 = {\rm Concat}(d_j, o_j)\mathbf{W}^x,
    \label{equ:get_node_feats}
\end{equation}
where we have $x_{i}^0, x_{j}^0 \in \mathbb{R}^{C_x}$. $\mathbf{W}^x \in \mathbb{R}^{(C+T)\times C_x}$ can be regarded as the weight of a linear layer. $T$ is the length of basic features for each object. $C_x$ is the encoded feature length of each node.
The edge feature $e_{ij}$ between $v_i$ and $v_j$ can be computed as
\begin{equation}
\label{equ:get_edge_feats}
    e_{ij}^0 = \mathcal{H}(x_j^0 - x_i^0, x_i^0),
\end{equation}
where $\mathcal{H}(\cdot)$ is a nonlinear transformation function that extracts the edge features. We prune the edges between the nodes which are more than $r$ meters away. $r$ is set to be 2 and the ablation studies on this aspect are presented in ~\secref{ab:radius}.
After getting all edge features, a channel-wise max-pooling aggregation is applied to each node $x_i$ to update each node's features:
\begin{equation}
\label{equ:gnn_aggregation}
    x_{i}^1 = \max_{j \in \vartheta} e_{ij}^0,
\end{equation}
where $\vartheta$ is the set of indices of nodes that are connected to the $i$-th node, 
$x_{i}^1$ indicates the refined features of the $i$-th node after one update,
and $\max$ is the channel-wise max-pooling operation on edge features.

Since the whole process can be iterative, we iteratively run the above operations (\equref{equ:get_edge_feats} - \equref{equ:gnn_aggregation}) for $m$ times to get features $x_i^m$ for each node. 
The final node features $x_i^{final}$ are the channel-wise concatenation of all hidden node features in each update:
\begin{equation}
    x_i^{final} = {\rm Concat}(x_i^1, x_i^2, ..., x_i^m).
\end{equation}

We feed  all final node features to the regression head that consists of three linear layers, predicting the refined locations, heading directions, and class labels.
The final loss of IntraRM can be defined as follows:
\begin{equation}
    \mathcal{L}_{intra} = \mathcal{L}_{cls} + \lambda_1 \mathcal{L}_{reg} + \lambda_2 \mathcal{L}_{dir},
\end{equation}
where $\mathcal{L}_{cls}$ is the focal loss \cite{lin2020focal}. $\mathcal{L}_{reg}$ and $\mathcal{L}_{dir}$ are the smooth L1 losses \cite{ren2016faster} for regressing object locations and heading directions, respectively.
$\lambda_1$ and $\lambda_2$ are the balance weights.

\begin{figure}
    \centering
    \includegraphics[width=8cm]{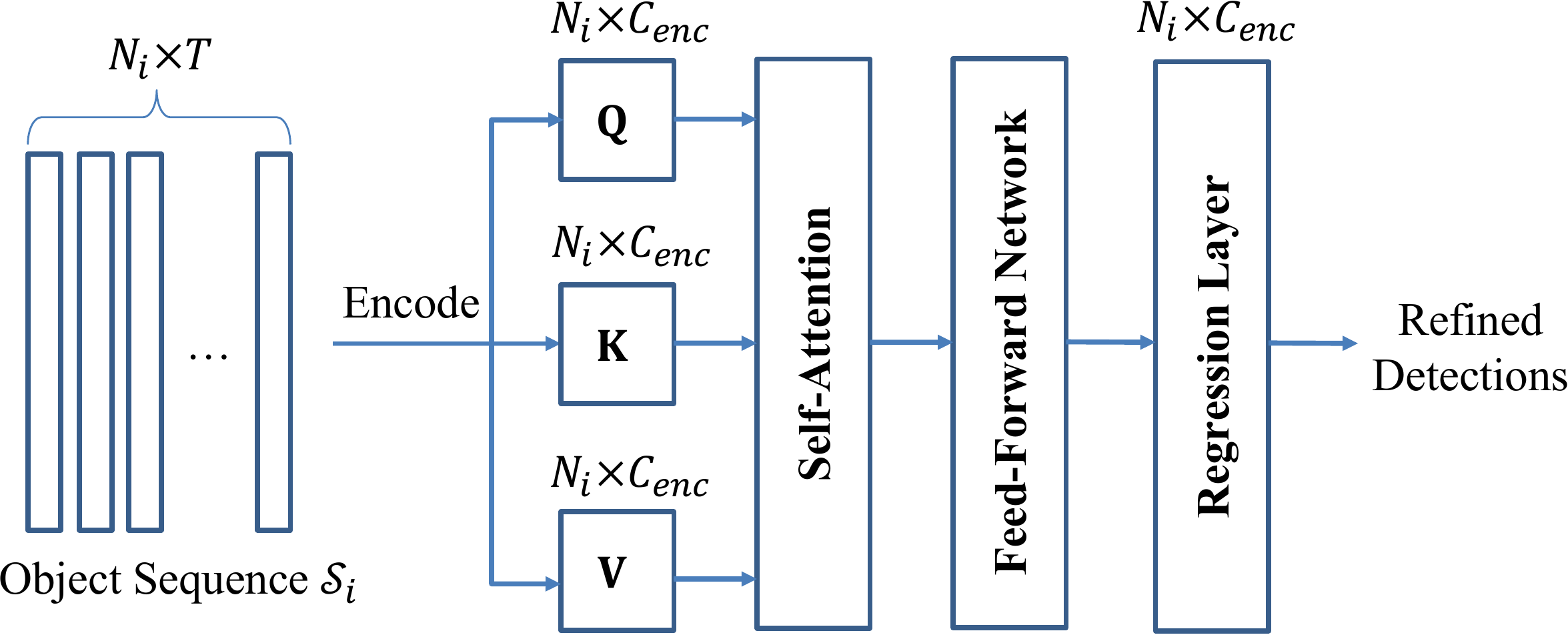}
    \vspace{-10pt}
    \caption{\textbf{The detailed structure of InterRM.}
    Given the tracked object sequence for each object, we perform transformer-based feature extraction for individual refinement.
    }
    \label{fig:interrm}
\end{figure}

\subsection{Inter-Frame Relation Module}
\label{sec:interrm}

Recently, there have been some works \cite{qi2021offboard,zakharov2020autolabeling} for offboard 3D object detection, 
which aims at improving the auto labeling accuracy from servers.
They aggregate point clouds from a long sequence for saturated performance,
which leads to heavy computational cost and is thus unsuitable for online applications like autonomous driving.
To address this issue, 
we propose a new efficient inter-frame relation module (InterRM).

Given a long LiDAR sequence, 
the one-stage base detector with IntraRM provides refined detection
results of each single LiDAR frame. On top of that, we further track each object via the tracker adopted in~\cite{yin2021center} 
and derive object sequences $\mathbb{S}=\{\mathcal{S}_1, \mathcal{S}_2\, ..., \mathcal{S}_n\}$.  As the same objects in the tracked sequences are highly correlated, a natural way is to model the inter-frame relationship with a densely connected graph. However, directly using message passing on such dense graph using GNNs brings significant extra overheads as discussed later. Instead, we resort to transformers due to its remarkable performance in modeling global relationships \cite{vaswani2017attention,dosovitskiy2021image}.

More specifically, for each object sequence $\mathcal{S}_i$ with the length of $N_i$, we have the sequence features 
$\{d_{i, -t_{N_i-1}}, d_{i, -t_{N_i-2}}, ..., d_{i, -t_1}, d_{i}  \}$  
predicted by the one-stage base detector,
where $\{-t_{N_i-1}, -t_{N_i-2}, ..., -t_1\}$ indicate the past time-stamps.
$N_i$ could be dynamic and is fully determined by the tracking results. 
In the object sequence $\mathcal{S}_i$,
the location features from past time-stamps
have been projected to the current frame's coordinate system.
We do channel-wise concatenation for the sequence features and encode the derived features using a linear layer:
\begin{equation}
\label{equ:inter_enc}
\mathbf{D} = {\rm Concat}(d_{i, -t_{N_i-1}}, d_{i, -t_{N_i-2}}, ..., d_{i, -t_1}, d_{i})^\mathbf{T} \mathbf{W}^e ,
\end{equation}
where $\mathbf{W}^e$ is the weight of the linear layer with $C_{enc}$ output channels,
$\mathbf{D} \in \mathbb{R}^{N_i\times C_{enc}}$ is the encoded features, 
and $N_i$ is the number of tracked frames for the $i$-th object.
Suppose that the query, key, and value in self-attention \cite{vaswani2017attention,dosovitskiy2021image} are represented as $\mathbf{Q}$, $\mathbf{K}$, and $\mathbf{V}$, which can be calculated as
\begin{equation}
(\mathbf{Q},\mathbf{K},\mathbf{V})=(\mathbf{D}\mathbf{W}^q,\mathbf{D}\mathbf{W}^k,\mathbf{D}\mathbf{W}^v),
\end{equation}
where $\mathbf{W}^q$, $\mathbf{W}^k$, and $\mathbf{W}^v$
indicate the weights of linear layers for computing query, key, and value, respectively.
Then we can compute the self-attention $\mathbf{A}$ for $\mathbf{D}$:
\begin{equation}
\label{equ:attention}
\mathbf{A} = {\rm Softmax}(\frac{\mathbf{Q}\times \mathbf{K}^\mathbf{T}}{\sqrt{C_K}})\times \mathbf{V},
\end{equation}
where $C_K$ is the number of channels of $\mathbf{K}$ and $\frac{1}{\sqrt{C_K}}$ is the scale factor of self-attention.
Note that the multiple head concept is omitted in \equref{equ:attention} for convenience.
A feed-forward network (FFN) learns the residual representation to strengthen the self-attention $\mathbf{A}$:
\begin{equation}
    \mathbf{A}' = {\rm FFN}(\mathbf{A} + \mathbf{D}) + \mathbf{A} + \mathbf{D},
\end{equation}
where ${\rm FFN}$ has two linear layers with an expansion rate of 2,
and $\mathbf{A}'$ is the refined feature.
Finally, we apply a regression layer to predict the locations and heading degrees. 
The smooth L1 loss \cite{ren2016faster} is selected as the regression loss.
It is also the
overall loss of InterRM $\mathcal{L}_{inter}$. The detailed structure of InterRM can be found in \figref{fig:interrm}.

\subsection{Time Complexity.}
\label{complexity}
Here, we analyze the time complexity of IntraRM and InterRM as below.

\myPara{IntraRM.}
Including the regressing head, the time complexity of IntraRM is $O(C_x^2n)$. Since each LiDAR frame usually has tens of objects, IntraRM only needs negligible computational cost (0.1G FLOPs).

\myPara{InterRM.}
We refine each object using its past tracked sequences.
For each object sequence, suppose the average length of object sequences is $N$,
the time complexity of the attention layer and linear layers is 
$O(C_{enc}N^2)$ and $O(C_{enc}^2N)$, respectively.
The overall time complexity of InterRM is $O(n(C_{enc}^2N+C_{enc}N^2))$.
Besides, on the dense graph with $\frac{N(N-1)}{2}$ edges, 
GNN with one iteration costs $O(nC_{enc}^2N^2)$, which practically results in 64.9G FLOPs, 20$\times$ more than using transformer in our setting.

\begin{table}[!t]
  \centering
  \caption{\textbf{Comparison with \sArt methods for vehicle detection.}
     Results are evaluated on the WOD validation set \cite{sun2020scalability}. 
     }
  \vspace{-10pt}
  \resizebox{\columnwidth}{!}{%
   \renewcommand\arraystretch{1.0}
   \setlength{\tabcolsep}{1mm}
    \begin{tabular}{c|lccc}
    \Xhline{1pt}
    Settings & Methods & Publication & mAPH  & mAP \\
    \Xhline{1pt}
    \multirow{16}[0]{*}{LEVEL\_1} & PointPillars \cite{lang2019pointpillars} & CVPR'19 & 62.8  & 63.3  \\
          & LaserNet \cite{meyer2019lasernet} & CVPR'19 & 50.1  & 52.1  \\
          & PV-RCNN \cite{shi2020pv} & CVPR'20 & - & 70.3 \\
          & PillarOD \cite{wang2020pillar} & ECCV'20 & -  & 69.8 \\
          & MVF \cite{zhou2020end}    & CoRL'20 & 62.9  & - \\
           & RCD \cite{bewley2020range}   & CoRL'20 & 69.6  & 69.2  \\
           & CVCNet \cite{chen2020every} & NeuIPS'20 & - & 65.2 \\
          & Voxel R-CNN \cite{deng2020voxel} & AAAI'21 & -     & 75.6  \\
          & PVGNet \cite{miao2021pvgnet} & CVPR'21 & -     & 74.0  \\
          & CenterPoint \cite{yin2021center} & CVPR'21 & 74.4  & 74.9  \\
          & LiDAR R-CNN \cite{li2021lidar} & CVPR'21 & 75.5  & 76.0  \\
          & RangeDet \cite{fan2021rangedet} & ICCV'21 & - & 72.9 \\
          & CT3D \cite{sheng2021improving}  & ICCV'21 & -     & 76.3  \\
          & VoTR-TSD \cite{mao2021voxel}  & ICCV'21 & 74.3  & 75.0  \\
          & Ret3D (Ours) & -     & \textbf{81.0} & \textbf{81.6} \\
    \Xhline{1pt}
    \multirow{9}[0]{*}{LEVEL\_2} 
    & PointPillars \cite{lang2019pointpillars} & CVPR'19 & 55.1  & 55.6  \\
          & PV-RCNN \cite{shi2020pv} & CVPR'20 & 63.7  & 64.2  \\
          & Voxel R-CNN \cite{deng2020voxel}  & AAAI'21 & -     & 66.6  \\
          & LiDAR R-CNN \cite{li2021lidar} & CVPR'21 & 67.9  & 68.3  \\
          & CenterPoint \cite{yin2021center} & CVPR'21 & 69.7  & 70.2  \\
          & VoTR-TSD \cite{mao2021voxel} & ICCV'21 & 65.3  & 65.9  \\
          & CT3D \cite{sheng2021improving}  & ICCV'21 & -     & 69.0  \\
          & Ret3D (Ours) & -     & \textbf{72.9} & \textbf{73.4} \\
    \Xhline{1pt}
    \end{tabular}%
    }
  \label{tab:cmp_sota_vehicle}%
  \vspace{-.5mm}
\end{table}%

\begin{table}[!t]
  \centering
  \caption{\textbf{Comparison with \sArt methods for pedestrian detection.} Results are evaluated on the WOD validation set \cite{sun2020scalability}. 
  }
  \vspace{-8pt}
  \resizebox{\columnwidth}{!}{%
   \renewcommand\arraystretch{1.0}
   \setlength{\tabcolsep}{1mm}
    \begin{tabular}{c|lccc}
    \Xhline{1pt}
    Settings & Methods & Publication & mAPH  & mAP \\
    \Xhline{1pt}
    \multirow{7}[0]{*}{LEVEL\_1} & PointPillars \cite{lang2019pointpillars} & CVPR'19 & 56.1  & 70.0  \\
    & PillarOD \cite{wang2020pillar} & ECCV'20 &  & 72.5 \\
    & MVF  \cite{zhou2020end}   & CoRL'20 & -     & 65.3  \\
          & PVGNet \cite{miao2021pvgnet} & CVPR'21 & -     & 69.5  \\
          & PointAugmenting \cite{wang2021pointaugmenting} & CVPR'21 & - & 75.4  \\
          & CenterPoint \cite{yin2021center} & CVPR'21 & 75.1  & 78.3  \\
          & RangeDet \cite{fan2021rangedet} & ICCV'21 & - & 75.9 \\
          & Ret3D (Ours) & -     & \textbf{79.7} & \textbf{82.8} \\
    \Xhline{1pt}
    \multirow{5}[0]{*}{LEVEL\_2} & PointPillars \cite{lang2019pointpillars} & CVPR'19 & 51.1  & 63.8  \\
          & PointAugmenting \cite{wang2021pointaugmenting} & CVPR'21 & -    & 70.6  \\
          & CenterPoint \cite{yin2021center} & CVPR'21 & 70.3  & 73.3  \\
          & Ret3D (Ours) & -     & \textbf{71.9} & \textbf{74.9} \\
    \Xhline{1pt}
    \end{tabular}%
  }
  \label{tab:sota_cmp_pedestrian}%
\end{table}%

\begin{table*}[!t]
  \centering
  \caption{\textbf{Effect of IntraRM and InterRM.}``*'' indicates that the computational cost is computed with 50 detected objects. 
Results are tested on the WOD validation set \cite{sun2020scalability} under the LEVEL\_2 setting.} 
  \vspace{-10pt}
  \resizebox{\textwidth}{!}{%
    \begin{tabular}{c|l|c|cc|cc|cc|cc}
    \Xhline{1pt}
    \multicolumn{1}{c|}{\multirow{2}[0]{*}{No.}} &
	\multicolumn{1}{c|}{\multirow{2}[0]{*}{Methods}} &
	\multicolumn{1}{c|}{\multirow{2}[0]{*}{\# FLOPs}} &
	\multicolumn{2}{c|}{\multirow{1}[0]{*}{Overall}}  &
	\multicolumn{2}{c|}{\multirow{1}[0]{*}{Vehicle}}  & 
	\multicolumn{2}{c|}{\multirow{1}[0]{*}{Pedestrian}}  & \multicolumn{2}{c}{\multirow{1}[0]{*}{Cyclist}}  \\
	& & & mAPH & mAP & mAPH & mAP & mAPH & mAP & mAPH & mAP \\ \Xhline{1pt}
	1 & CenterPoint \cite{yin2021center} & 127.7G & 68.2  & 69.9  & 67.3  & 67.8  & 67.5  & 71.0    & 69.9  & 70.8   \\ 
	2 & No.1 + Two-stage \cite{yin2021center} & 0.1G* & 70.3  & 71.7  & 69.7  & 70.2  & 70.3  & 73.3  & 70.9  & 71.7 \\  
	3 & No.1 + IntraRM (Ours)  & 0.1G* & 71.1  & 72.5  & 70.9  & 71.4  & 70.9  & 73.8  & 71.6  & 72.4  \\ 
	4 & No.3 + InterRM (Ours) & 3.2G* &\textbf{72.3} & \textbf{73.8} & \textbf{72.9} & \textbf{73.4} & \textbf{71.9} & \textbf{74.9} & \textbf{72.2} & \textbf{73.0} \\
	- & \textit{Improvement}  & - & \highlight{+4.1} & \highlight{+3.9} & \highlight{+5.6} & \highlight{+5.6} 
	& \highlight{+4.4} & \highlight{+3.9}
	& \highlight{+2.3} & \highlight{+2.2} \\
	\hline
	5 & SECOND \cite{yan2018second}  & 124.0G & 59.3 & 64.7 & 65.1  & 65.7 & 53.5  & 62.8    & 59.3 & 65.6   \\ 
	6 & No.5 + IntraRM (Ours)  & 0.1G* & 63.6 & 67.0 & 66.8 & 67.4 & 57.6 & 65.9 & 66.4 & 67.6  \\
	7 & No.6 + InterRM (Ours)  & 3.2G* & 64.8 & 68.2 & 69.2 & 69.8 & 58.6 & 66.8 & 66.7 & 67.9 \\
	- & \textit{Improvement}  & - & \highlight{+5.5} & \highlight{+3.5} & \highlight{+4.1} & \highlight{+4.1} 
	& \highlight{+5.1} & \highlight{+4.0}
	& \highlight{+7.4} & \highlight{+2.3} \\
	\Xhline{1pt}
\end{tabular}}
  \label{tab:centerpoint_cmp}
\end{table*}

\section{Experiments}

\subsection{Experimental Setup}

\myPara{Implementation details.}
We implement the proposed Ret3D framework using the PyTorch library \cite{paszke2019pytorch}.
We train our networks using the AdamW \cite{loshchilov2017decoupled} optimizer, with the weight decay of 0.01.
Two recent popular one-stage universal detectors, SECOND \cite{yan2018second} and CenterPoint \cite{yin2021center},
are selected as our base detectors.
Both detectors use sparse 3D convolutions as the backbone for feature extraction,
with the voxel size of $\{0.1m, 0.1m, 0.15m\}$.
We reproduce SECOND \cite{yan2018second} with OpenPCDet \cite{openpcdet2020}, a well-known open-source toolbox. For CenterPoint \cite{yin2021center}, we directly use the pretrained models provided by the authors.
For IntraRM, we use a learning rate of $3\times 10^{-4}$ with 16 LiDAR frames per mini-batch.
For InterRM, we  apply the same learning rate with 64 object sequences per mini-batch.
We train IntraRM and InterRM for 100K iterations, respectively.
The loss balance weights $\lambda_1$ and $\lambda_2$ are empirically set to 2.0 and 0.2, respectively.
We adpot the EdgeConv \cite{wang2019dynamic} as the nonlinear transformation function of \equref{equ:get_edge_feats}.
We use \cite{yin2021center} as the base tracker for InterRM,  in which the average object sequence length $N$ is 100.
$C_{enc}$ is set to 256.
We use 16 heads in computing $\mathbf{A}$ using \equref{equ:attention} for catching diverse attention.

\myPara{Dataset.}
We apply the recently proposed Waymo Open Dataset (WOD) \cite{sun2020scalability}
for training and evaluation.
Unlike the KITTI \cite{geiger2012we} dataset that only annotates 15K LiDAR frames in total,
WOD has 798 training sequences and 202 validation sequences with 158K and 40K annotated LiDAR frames, respectively.
Each sequence is sampled with a frequency of 10Hz.
WOD has 12M 3D boxes, 150$\times$ more than that in the KITTI dataset \cite{geiger2012we}. 
WOD is also much more challenging because it contains more complicated scenes.
We train the proposed Ret3D on the training set, and 
the validation set is used for evaluation.
We compare our framework with recent \sArt methods on the full validation set.
The range of detection results is $[-75.2m, 75.2m]$ for x and y axes, and $[-2m, 4m]$ for the z axis, so we have $41\times 1504 \times 1504$ voxels as the input for the base detector.

\myPara{Evaluation metric.}
Following the advice of WOD \cite{sun2020scalability},
mAPH, \ie, average precision (AP) weighted by the heading accuracy, is the primary metric.
More details about mAPH can refer to the original paper \cite{sun2020scalability}.
We also report mAP for reference.
As suggested by WOD \cite{sun2020scalability}, the IoU thresholds of vehicle, pedestrian, and cyclist categories are set to 0.7, 0.5, and 0.5, respectively.
The evaluation results under both LEVEL\_2 (more difficult, including 3D boxes that contain less than five LiDAR points) and LEVEL\_1 settings will be reported.

\renewcommand{\AddImg}[1]{
  \includegraphics[width=0.32\textwidth, height=0.24\textwidth]{Imgs/#1}
}

\begin{table}[!t]
  \centering
  \caption{\textbf{Positional encoding in InterRM.} ``PE'' denotes the positional encoding. Results are tested on the WOD validation set \cite{sun2020scalability} under the LEVEL\_2 setting.}
  \vspace{-10pt}
  \resizebox{\linewidth}{!}{%
    \begin{tabular}{l|cc|cc|cc}
    \Xhline{1pt}
	\multicolumn{1}{c|}{\multirow{2}[0]{*}{PE Settings}} & \multicolumn{2}{c|}{\multirow{1}[0]{*}{Vehicle}}  & \multicolumn{2}{c|}{\multirow{1}[0]{*}{Pedestrian}}  & \multicolumn{2}{c}{\multirow{1}[0]{*}{Cyclist}}  \\
	 & mAPH & mAP & mAPH & mAP & mAPH & mAP \\
	 \Xhline{1pt}
	Baseline & 70.9 & 71.4 & 70.9 & 73.8 & 71.6 & 72.4 \\
	+ Implicit & \textbf{72.9} & \textbf{73.4} & \textbf{71.9} & \textbf{74.9} & 72.2 & 73.0 \\ 
	++Temporal & 72.7 & 73.1 & 71.8 & 74.7 & \textbf{72.6} & \textbf{73.4} \\ 
	++Spatial & 71.9 & 72.4 & 71.4 & 74.3 & 72.0 & 72.7 \\
	\Xhline{1pt}
\end{tabular}}
  \label{tab:inter_pe}
\end{table}

\subsection{Evaluation Results}

\myPara{Vehicle detection.}
We compare the proposed Ret3D framework with 14 \sArt methods published in the recent three years. 
Results of these methods are reported by their official papers or reproduced using the official pretrained models.
We show the comparison results in \tabref{tab:cmp_sota_vehicle}.
Under the LEVEL\_1 setting, we achieve 5.5\% and 5.3\% improvement compared with the best competitor in terms of mAPH and mAP, respectively.
Under the more difficult LEVEL\_2 setting, we still obtain 3.2\% gain in terms of both mAPH and mAP.
Such significant improvement shows that 
object relations are very helpful for vehicle detection.

\myPara{Pedestrian detection.}
Pedestrian detection is also very important for autonomous driving.
We still compare our Ret3D with \sArt methods in the recent three years that have reported the results for pedestrian detection.
The comparison results are shown in \tabref{tab:sota_cmp_pedestrian}.
Ret3D achieves 4.6\% and 1.6\% improvement in terms of mAPH under LEVEL\_1 and LEVEL\_2, respectively.
This demonstrates the superiority of Ret3D in pedestrian detection.

\subsection{Ablation Study}

\myPara{Effect of IntraRM and InterRM.}
As described in \secref{sec:intro}, we explore how to explicitly leverage 
intra-frame and inter-frame object relations for improving the performance of modern efficient 3D object detectors.
To this end,
we propose Ret3D which consists of two modules, IntraRM and InterRM.
To validate the effectiveness of the proposed IntraRM and InterRM,
we conduct the ablation experiments in \tabref{tab:centerpoint_cmp}.
We take the general center-based detector CenterPoint \cite{yin2021center} and anchor-based detector SECOND \cite{yan2018second}  as the base detectors.
We also compare our framework with two-stage CenterPoint \cite{yin2021center} that refines each detection result individually.
We can observe that IntraRM has a significant improvement over the base detector (+2.9\% in mAPH and +2.6\% in mAP), 
demonstrating the superiority of leveraging intra-frame object relations.
Replacing two-stage CenterPoint with IntraRM leads to 0.8\% mAPH and mAP gain under the LEVEL\_2 setting, 
implying that intra-frame relations can even be more significant than independent refinement of each object.
Based on the base detector with IntraRM (No.3),
adding InterRM can further improve the performance significantly (+1.2\% in mAPH and +1.3\% in mAP).
Overall, Ret3D can achieve 4.1\% mAPH and 3.9\% mAP improvement over the CenterPoint \cite{yin2021center} under the LEVEL\_2 setting, respectively.
No.5 - 7 results of \tabref{tab:centerpoint_cmp} also show that Ret3D has +5.6\% mAPH and 3.5\% mAP improvement on the SECOND \cite{yan2018second} framework.
The above analyses suggest that both intra-frame and inter-frame object relations are very significant 
for improving efficient 3D object detection performance for autonomous driving.

For efficiency analysis, we report the number of FLOPs in \tabref{tab:centerpoint_cmp}. 
We can see that both base detectors cost more than 100G FLOPs (No. 1, 5).
Two-stage CenterPoint \cite{yin2021center} (No. 2) introduces extra small network complexity and negligible computational cost of 0.1G FLOPs, 
as it only does individual refinement based on the BEV feature vectors.
IntraRM introduces comparable computational cost with two-stage CenterPoint \cite{yin2021center}, 
but InterRM requires more computational cost. 
Nevertheless, InterRM still has much less computational cost than 
base detectors, keeping high efficiency for 3D object detection.

\myPara{Positional encoding in InterRM.}
Generally, the transformer \cite{vaswani2017attention,dosovitskiy2021image} prefers the input with positional encoding \cite{carion2020end}.
By default, we use a linear layer to encode the detection results (\equref{equ:inter_enc}), 
which can be viewed as an implicit way of temporal and spatial positional encoding.
Here, we externally add two types of explicit positional encoding, \ie, temporal or spatial encoding, by directly adding temporal or spatial terms on the encoded features $\mathbf{D}$.
The results are shown in \tabref{tab:inter_pe}.
The baseline is CenterPoint \cite{yin2021center} with IntraRM.
We find that using temporal encoding will slightly decrease the performance of vehicle and pedestrian classes
while giving slightly better performance for the cyclist class.
Spatial encoding will significantly downgrade the detection accuracy of all three classes.
Therefore, we do not add explicit positional encoding in practice.

\begin{table}[!t]
  \centering
  \caption{\textbf{The radius setting in IntraRM.} Results are tested on the WOD validation set \cite{sun2020scalability} under the LEVEL\_2 setting.}
  \vspace{-10pt}
  \resizebox{.47\textwidth}{!}{%
    \begin{tabular}{c|cc|cc|cc}
    \Xhline{1pt}
	\multicolumn{1}{c|}{\multirow{2}[0]{*}{Radius}} & \multicolumn{2}{c|}{\multirow{1}[0]{*}{Vehicle}}  & \multicolumn{2}{c|}{\multirow{1}[0]{*}{Pedestrian}}  & \multicolumn{2}{c}{\multirow{1}[0]{*}{Cyclist}}  \\
	& mAPH & mAP & mAPH & mAP & mAPH & mAP \\ \Xhline{1pt}
	- & 65.1 & 65.7 & 53.5 & 62.8 & 59.3 & 65.6 \\
	1.0m & 65.9 & 66.5 & 57.6 & 65.7 & 65.1 & 66.4 \\
	2.0m & \textbf{66.8} & \textbf{67.4} & \textbf{57.6} & \textbf{65.9} & \textbf{66.4} & \textbf{67.6} \\
	4.0m & \textbf{66.8} & \textbf{67.4} & \textbf{57.6} & 65.8 & 66.3 & 67.5 \\
	\Xhline{1pt}
\end{tabular}}

  \label{tab:radius}
\end{table}

\begin{table}[!t]
  \centering
  \caption{\textbf{Number of iterations $m$ for graph update.} Results are tested on the WOD validation set \cite{sun2020scalability} under the LEVEL\_2 setting.}
  \vspace{-10pt}
  \resizebox{.47\textwidth}{!}{%
    \begin{tabular}{c|cc|cc|cc}
    \Xhline{1pt}
	\multicolumn{1}{c|}{\multirow{2}[0]{*}{\# Iters}} & \multicolumn{2}{c|}{\multirow{1}[0]{*}{Vehicle}}  & \multicolumn{2}{c|}{\multirow{1}[0]{*}{Pedestrian}}  & \multicolumn{2}{c}{\multirow{1}[0]{*}{Cyclist}}  \\
	 & mAPH & mAP & mAPH & mAP & mAPH & mAP \\ \hline
	 	- & 65.1 & 65.7 & 53.5 & 62.8 & 59.3 & 65.6 \\
	1 & 66.6 & 67.2 & 57.4 & 65.6 & 65.6 & 66.9 \\ 
	2 & 66.7 & 67.3 & 57.4 & 65.7 & 65.9 & 67.1 \\ 
	4 & \textbf{66.8} & \textbf{67.4} & \textbf{57.6} & \textbf{65.9} & \textbf{66.4} & \textbf{67.6} \\ 
	6 & \textbf{66.8} & \textbf{67.4} & 57.5 & \textbf{65.9} & 66.3 & 67.5 \\ \Xhline{1pt}
\end{tabular}}
  \label{tab:intra_iters}
\end{table}

\myPara{The radius $r$ in constructing graph.}
\label{ab:radius}
In IntraRM, we construct the graph whose edges only connect adjacent node pairs.
For each node $v_i$, we only connect nodes whose spatial centers 
are within the circle centered with the center of node $v_i$ with the radius of $r$ meters.
The larger $r$ is, the more dense the constructed graph is.
Larger $r$ is expected to achieve better performance ideally, while it introduces more computational cost due to the more complex graph.
Based on our observation, we perform experiments using different $r$ in IntraRM.
The results are displayed in \tabref{tab:radius}. We use
SECOND \cite{yan2018second} as the baseline.
We find that $r\ge 1.0$m is enough for pedestrian and vehicle detection.
The larger $r$ is, the better performance of cyclist detection is.
However, we find that $r=4$m has slightly worse performance and
produces $3\times$ more edges in the graph.
Considering the trade-off between efficacy and efficiency, we apply $r=2$m.

\myPara{The number of iterations $m$ for graph update.}
As the graph update process can be iterative, we can iteratively refine the node features in the graph.
We show the effect of running the different number of iterations in \tabref{tab:intra_iters}.
SECOND \cite{yan2018second} is used as the baseline.
We can observe that different settings of $m$ only slightly affect the performance of vehicle and pedestrian detection. 
$m=4$ achieves the best result in vehicle detection and the second-best in pedestrian detection. 
For cyclist detection, $m=4$ achieves the best result.
Therefore, we apply $m=4$ for graph update.

\section{Conclusion}

To explicitly leverage intra-frame and inter-frame object relations
for improving 3D object detection,
we propose a simple, efficient, and effective framework called Ret3D.
Ret3D utilizes a two-stage pipeline and refines the detection results from one-stage detectors using intra-frame and inter-frame relation modules.
The experimental results demonstrate the effectiveness of the above two types of object relations.
Benefited by the above two types of object relations, on the WOD validation set \cite{sun2020scalability},
Ret3D achieves the new \sArt performance,
surpassing 5.5\% and 3.2\% in terms of the LEVEL\_1 and LEVEL\_2 mAPH 
compared with recent popular methods.

{\small
\bibliographystyle{IEEEtran}
\bibliography{references}
}

\end{document}